\documentclass{article}
\usepackage{spconf,amsmath,graphicx,hyperref}

\title{Leveraging audio-visual data to reduce the multilingual gap in self-supervised speech models}
\name{Mar\'ia Andrea Cruz Bland\'on$^{\dagger\star}$\thanks{$^{\star}$ Work done while interning at Apple} \qquad Zakaria Aldeneh$^{\ddagger}$ \qquad Jie Chi$^{\ddagger}$ \qquad Maureen de Seyssel$^{\ddagger}$}
  \address{$^{\dagger}$ Tampere University \qquad $^{\ddagger}$ Apple}

\usepackage[table]{xcolor}

\usepackage{multirow}

\begin{document}
%\ninept
%
\maketitle
\begin{abstract}
Self-supervised learning (SSL) has made significant advances in speech representation learning. Models like wav2vec 2.0 and HuBERT have achieved state-of-the-art results in tasks such as speech recognition, particularly in monolingual settings. However, multilingual SSL models tend to underperform their monolingual counterparts on each individual language, especially in multilingual scenarios with few languages such as the bilingual setting. In this work, we investigate a novel approach to reduce this performance gap by introducing limited visual grounding into bilingual speech SSL models. 
Our results show that visual grounding benefits both monolingual and bilingual models, with especially pronounced gains for the latter, reducing the multilingual performance gap on zero-shot phonetic discrimination from 31.5\% for audio-only models to 8.04\% with grounding.
% Our results show that visual grounding substantially reduces the performance gap between monolingual and bilingual models on zero-shot phonetic discrimination, from 31.5\% relative difference to 8.04\%.
% While both monolingual and bilingual models benefit from visual grounding, the gains are especially pronounced for bilingual models, highlighting its effectiveness in reducing the multilingual gap.

\end{abstract}
\section{Introduction}
\label{sec:intro}

Self-supervised learning (SSL) has made significant advances in speech representation learning. Speech representations from models like wav2vec 2.0 \cite{Baevski2020} and HuBERT \cite{Hsu2021} have achieved state-of-the-art results in downstream tasks such as automatic speech recognition (ASR), particularly in monolingual settings. More interestingly, these models have shown potential to support pipelines for textless applications \cite{lakhotia2021}, enabling direct processing of speech without transcription. However, multilingual SSL models typically exhibit a ``multilingual gap'', i.e., they underperform their monolingual counterparts on each individual language, in both zero-shot and downstream tasks \cite{deseyssel2023, lehecka24_interspeech}. This behavior suggests that learning speech representations from multiple languages simultaneously creates interference between language-specific linguistic patterns, degrading representation quality for each individual language. While this multilingual gap can apparently be overcome with large multilingual training (over 50 languages and tens of thousands of hours; \cite{conneau21_interspeech}), such an approach is computationally expensive and data-intensive. This challenge raises the question of whether more data-efficient solutions exist to address multilingual interference. 

Research in child language development offers insights into this challenge. Bilingual infants develop language abilities comparably to monolingual infants despite having reduced exposure to each individual language compared to their monolingual peers \cite{Hohle2020}, suggesting that efficient learning mechanisms can overcome limited per-language exposure. In fact, prior research has suggested that bilingual infants may employ slightly different learning mechanisms as there is evidence that they pay more attention to visual cues compared to monolingual infants \cite{DSOUZA2021727}. Drawing inspiration from this observation, \textit{we explore whether multimodal approaches, particularly visual grounding, help computational models reduce the multilingual gap more efficiently}.  Our hypothesis draws from the idea that visual information could serve as an additional signal to help models better discriminate between languages during training, potentially reducing cross-lingual interference and leading to more robust language-specific representations.

Prior work has demonstrated the benefits of visual grounding in speech SSL. In monolingual settings, visually grounded speech (VGS) models can improve phonetic discrimination in zero-shot settings \cite{Khorrami2023}, enable alignment of speech segments with visual context \cite{Shih2023}, and yield more noise-robust representations than audio-only SSL models \cite{shi2022}. These benefits also extend to multilingual settings. Evidence suggests visual modality can function as an interlingual bridge, sometimes enabling multilingual VGS models to outperform their monolingual VGS counterparts in retrieval task \cite{Harwath2018}. Multilingual VGS models---whether using language-specific encoders \cite{Harwath2018, Ohishi2020} or more scalable language-agnostic approaches \cite{Berry2023, han-etal-2024-xlavs}---achieved better downstream performance in tasks such as ASR than audio-only models. However, previous work has not specifically examined whether visual grounding can address the fundamental multilingual gap in SSL pretraining. 

In this work, we directly investigate whether limited visual grounding via image-spoken caption pairs during pretraining can reduce interference between languages and help reduce the multilingual gap. To our knowledge, this is the first work directly testing visual grounding as a solution to this gap in SSL speech models.

\section{Proposed Method}
\label{sec:methodology}
\subsection{Hypotheses}

We test whether limited visual grounding can reduce the performance gap between monolingual and bilingual models in zero-shot evaluation. To this end, we design controlled experiments around two hypotheses:

\begin{figure}
    \centering
    \includegraphics[width=1\linewidth]{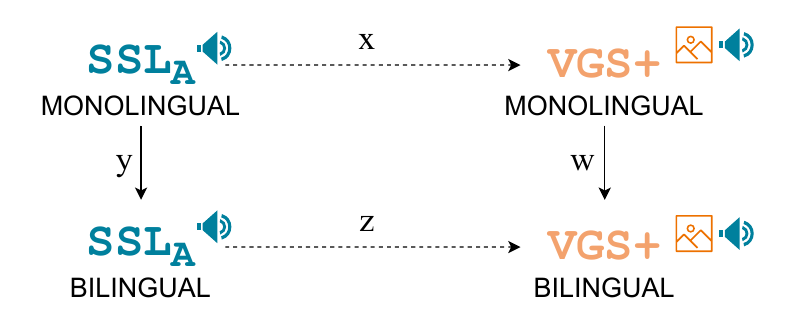}
    \caption{
    We compare two quantities: (1) the monolingual–bilingual performance gap reduction in audio-only (SSL$_A$, gap = $y$) vs. visually grounded (VGS$+$, gap = $w$) models, testing whether $y > w$;  and (2) the visual grounding gains for monolingual (gain = $x$) vs. bilingual (gain = $z$) models, testing whether $z > x$.}

    \label{fig:method}
\end{figure}

 \vspace{0.5em}
 
\noindent \textbf{(1) Gap reduction.} Visual grounding \emph{reduces the multilingual gap}. In Fig.~\ref{fig:method}, we compare the performance difference between monolingual and bilingual models in audio-only settings (gap = $y$) with the same comparison in visually grounded models (gap = $w$). We test whether $y > w$, i.e., whether adding visual grounding reduces the gap.

 \vspace{0.5em}

\noindent \textbf{(2) Differential benefit.} Visual grounding is \emph{more beneficial for multilingual} than monolingual models. In Fig.~\ref{fig:method}, we compare the gains from adding visual grounding in monolingual models (gain = $x$) with the improvement in bilingual models (gain = $z$). We test whether $z > x$, i.e., whether multilingual models gain more from visual grounding.

 \vspace{0.5em}

If both hypotheses hold, this would suggest that visual information serves as an effective interlingual bridge that specifically helps multilingual models overcome cross-lingual interference.

\subsection{Setup}
Following the findings of \cite{Khorrami2023, han-etal-2024-xlavs}, we adopt a sequential training schedule where models are first trained on audio-only data before introducing audio-visual paired data (hereafter \texttt{VGS+}). To isolate the impact of visual grounding, we create control models that follow an identical training schedule but without the visual information, i.e., first training on the same audio-only data, followed by the audio part of the audio-visual paired data (hereafter \texttt{SSL$_{A}$}). With this design, we account for possible discrepancies in performance due to possible mismatch in the distributions of the training data. 

Moreover, to separate multilingual interference effects from data quantity effects, we make the total training hours equal across all models. Bilingual models receive half the per-language data of monolingual models, maintaining the same total hours (e.g., 1000h total: either all EN/FR for monolingual or 500h EN + 500h FR for bilingual setup). This setup directly tests whether the multilingual gap persists even with equal computational/data budgets, which is central to our research question.

\subsection{Datasets}
\label{sec:datasets}

First, we train our models on audio-only data using two read speech corpora (audiobooks): LibriLight (English) \cite{kahn2020libri} and Audiocite (French) \cite{felice2024audiocite}. Each corpus is randomly subsampled to about 1,000 hours of audio, with a matched number of speakers and balanced gender distribution.

For the second stage of training, we use the cross-modal-3600 \cite{ThapliyalCrossmodal2022} corpus, which comprises a machine translated version of the MS-COCO image–caption dataset \cite{Lin2014} in 34 languages. The selected subset (hereafter ML-COCO) contains 288k captions for training and 25k for validation\footnote{We use about 50\% of the original training split in \cite{ThapliyalCrossmodal2022} due to computational constraints for the synthesis}. We synthesize English and French captions with internal text-to-speech tools, yielding about 275 hours of speech per language. For each language, we generate speech with one male and one female voice, maintaining gender balance across captions. For the image side, we rely on the precomputed embeddings provided by LXMERT \cite{tan-bansal-2019-lxmert}, extracted as Faster RCNN region of interest (RoI) features from the original MS-COCO.

For our tests, we use the data partition from Common Voice \cite{ardila-etal-2020-common} English and French used in \cite{lavechin2025simulating, deseyssel2023}. 

\subsection{Models}

\begin{figure}
    \centering
    \includegraphics[width=1\linewidth]{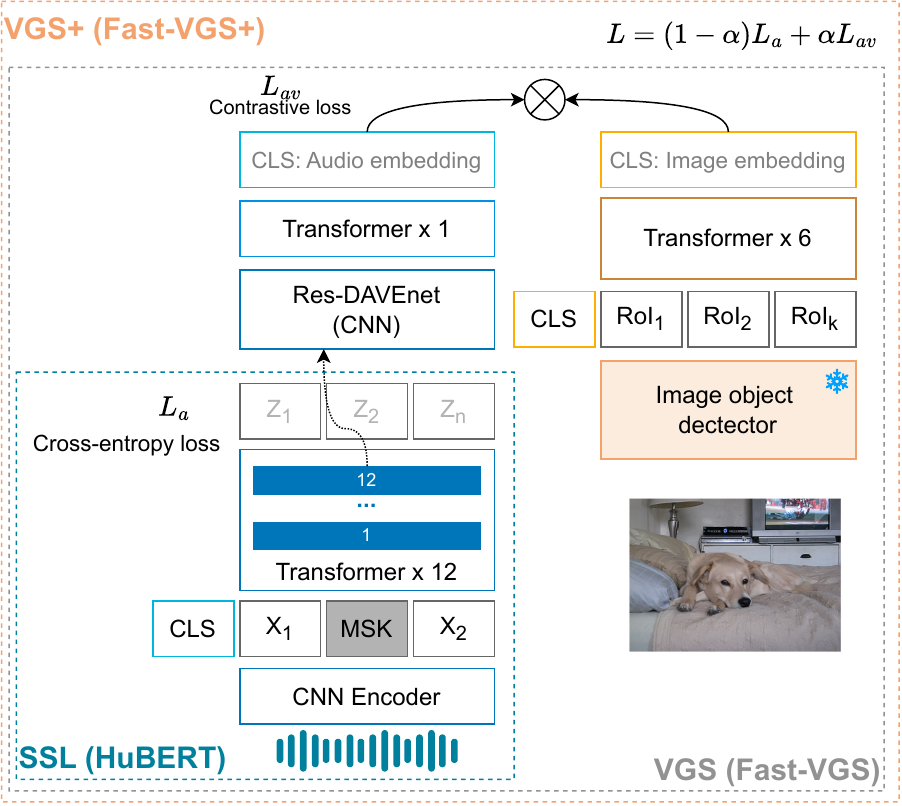}
    \caption{FaST-VGS+ architecture \cite{peng2022, Khorrami2023} using HuBERT-base \cite{Hsu2021} as the audio encoder.}
    \label{fig:model_arch}
\end{figure}

As the audio-only SSL model, we use HuBERT-base \cite{Hsu2021}, trained with a masked token prediction objective. Specifically, the model learns to classify pseudo-labels obtained by clustering input representations in a two-stage process, optimized with a cross-entropy loss ($L_{a}$).

As the visually-grounded SSL model, we extend the HuBERT-base model using the FaST-VGS+ architecture \cite{peng2022, Khorrami2023} (see Fig.~\ref{fig:model_arch}). While the original FaST-VGS+ used the Wav2Vec 2.0 model as its audio encoder, we successfully replicate the performance pattern reported in \cite{Khorrami2023} with HuBERT-base (experiments in \cite{Khorrami2023} were carried out using Wav2Vec 2.0). We use the last transformer layer of HuBERT as input to the Res-DAVEnet block, which is followed by one transformer layer. For the image branch, we use a 6-layer transformer block that processes RoI features. During VGS+ training, we freeze the CNN encoder of the audio model as in \cite{Khorrami2023}.

FaST-VGS+ is optimized using two losses, the audio loss ($L_{a}$) and the audio-visual alignment loss ($L_{av}$). The latter corresponds to a contrastive loss in which the positive sample is the caption-image pair and the negative samples are derived from permutation of all the other pairs in the batch. Similarly to \cite{Khorrami2023}, we use equal weights for the $L_{a}$ and the $L_{av}$, i.e., $\alpha=0.5$ in the final loss ($ L = (1 - \alpha) L_{a} + \alpha L_{av}$). 

Following \cite{han-etal-2024-xlavs}, to reduce the number of iterations required to train the HuBERT model when including audio-visual data, we use the same $k$-means model derived from the first stage of audio-only training for processing the pseudo-labels for the audio subset of the audio-visual data. 

We use a learning rate of 0.0005 for the first stage of the training (i.e., audio-only with the read speech corpora, \texttt{SSL}) and train the model up to 90k update steps (about 60 epochs). For the second stage, training with ML-COCO corpus (\texttt{SSL$_{A}$} or \texttt{VGS+}), we use a learning rate of 0.0001 and train for up to 2k update steps (about 5 epochs). We optimize the model using AdamW optimizer. Training was carried out using 32 A100 GPUs with 87.5 seconds per batch. 

\subsection{Evaluations}
\label{sec:test_taks}

We evaluate the linguistic content captured in the latent representations of the audio encoder of both audio-only and visually-grounded models in a zero-shot setup. We choose two tasks that directly assess the core linguistic capabilities affected by the multilingual gap: phonetic discrimination (the main focus of our study) and language identification (to understand cross-lingual interference patterns).

We extract the latent representations from the last transformer layer (layer 12) of the audio encoder (SSL HuBERT in Fig.~\ref{fig:model_arch}) and assess the phonetic learning using the ABX phoneme discrimination task \cite{schatz2016}. 
In the ABX task, the model is given two exemplars from different classes (A and B) and a third exemplar (X) that belongs to the same class as A. The task measures the distances between A–X and B–X in the embedding space. If the model captures class information, the A–X distance should be smaller than the B–X distance.
To further investigate the content in the speech representations, we also test the models with the ABX language discrimination task \cite{deseyssel2023}. 

For both ABX tasks, we use cosine distance as our similarity metric. For phonetic discrimination, we apply the Dynamic Time Warping (DTW) algorithm to align sequences, while for language discrimination, we represent each utterance by mean-pooling across all frames. 
ABX phonetic discrimination scores are computed under two conditions: comparing samples from the same speaker (within-speaker) and from different speakers (across-speaker). For the calculation itself, we employ fastABX \cite{poli2025fastabx}.

\section{Results}
\label{sec:result}

Table~\ref{tab:results} reports ABX error rates for phoneme discrimination (within- and across-speaker) in the native setup, i.e., testing each model on the same language on which it was trained.  
We focus on the \textbf{average performance} of monolingual models (EN and FR) versus bilingual models, with performance differences denoted $x$, $y$, $z$, and $w$ as in Fig.~\ref{fig:method}.

\noindent \textbf{Multilingual gap.} The first block of the table shows ABX phonetic discrimination error rates after training on read speech (\texttt{SSL}). When comparing monolingual and bilingual models trained on the same amount of data (1k hours), we observe a clear multilingual gap (monolingual: $6.28\%$ within-speaker, $7.5\%$ across-speaker; bilingual: $7.1\%$ within-speaker, $8.78\%$ across-speaker). This gap persists, though less prominently, even when bilingual models are trained with twice as much data (\textit{bilingual 2k}), consistent with prior findings \cite{deseyssel2023, lehecka24_interspeech}. \\

\begin{table}[t]
\small
\centering
\resizebox{\columnwidth}{!}{%
\begin{tabular}{ll l r r r}
\hline
\multirow{2}{*}{} & \multirow{2}{*}{Language} & \multirow{2}{*}{Dataset}
  & \multicolumn{3}{c}{ABX phonetic disc. (\%) $\downarrow$} \\
& & & WS & AS & \textbf{Avg} \\ \hline

\multirow{6}{*}{\rotatebox[origin=c]{90}{\texttt{SSL}}}
  & Monolingual 1k & LL/AC & 6.28 & 7.50 & \textbf{6.89} \\
  & \multicolumn{1}{r}{\color{gray} EN} & {\color{gray} LL}
      & {\color{gray} 6.63} & {\color{gray} 7.94} & {\color{gray}\textbf{7.29}} \\
  & \multicolumn{1}{r}{\color{gray} FR} & {\color{gray} AC}
      & {\color{gray} 5.93} & {\color{gray} 7.05} & {\color{gray}\textbf{6.49}} \\
  & Bilingual 1k & LL\&AC & 7.10 & 8.78 & \textbf{7.94} \\
  & Bilingual 2k & LL\&AC & 6.45 & 7.82 & \textbf{7.14} \\
\hline

\multirow{5}{*}{\rotatebox[origin=c]{90}{\texttt{SSL$_A$}}}
  & Monolingual & LL/AC+ML & 6.46 & 7.75 & \textbf{7.11} \\
  & \multicolumn{1}{r}{\color{gray} EN} & {\color{gray} LL+ML(EN)}
      & {\color{gray} 6.64} & {\color{gray} 7.90} & {\color{gray}\textbf{7.27}} \\
  & \multicolumn{1}{r}{\color{gray} FR} & {\color{gray} LL+ML(FR)}
      & {\color{gray} 6.29} & {\color{gray} 7.60} & {\color{gray}\textbf{6.95}} \\
  & Bilingual 1k & LL\&AC+ML & 8.36 & 10.34 & \textbf{9.35} \\
  \rowcolor[gray]{.9}
  & \multicolumn{2}{r}{\textit{Multilingual gap $y$ (relative \%)}} 
      & \textit{29.41} & \textit{33.42} & \textit{\textbf{31.5}} \\
\hline

\multirow{5}{*}{\rotatebox[origin=c]{90}{\texttt{VGS+}}}
  & Monolingual & LL/AC+ML & 5.86 & 6.81 & \textbf{6.34} \\
  & \multicolumn{1}{r}{\color{gray} EN} & {\color{gray} LL+ML(EN)}
      & {\color{gray} 6.30} & {\color{gray} 7.25} & {\color{gray}\textbf{6.78}} \\
  & \multicolumn{1}{r}{\color{gray} FR} & {\color{gray} AC+ML(FR)}
      & {\color{gray} 5.43} & {\color{gray} 6.37} & {\color{gray}\textbf{5.90}} \\
  & Bilingual 1k & LL\&AC+ML & 6.18 & 7.52 & \textbf{6.85} \\
  \rowcolor[gray]{.9}
  & \multicolumn{2}{r}{\textit{Multilingual gap $w$ (relative \%)}} 
      & \textit{5.46} & \textit{10.43} & \textit{\textbf{8.04}} \\
\hline
\end{tabular}
}
\caption{ABX phonetic discrimination error rates. EN and FR subrows (gray) use their own native test sets; all other rows use the combined EN+FR test set for comparability. Shaded rows report the multilingual gap (relative \%, i.e., $y$ for \texttt{SSL$_A$}, $w$ for \texttt{VGS+}). WS: within-speaker, AS: across-speaker, Avg: average. LL: LibriLight, AC: Audiocite, ML: ML-COCO.}
\label{tab:results}
\end{table}

\noindent \textbf{(1) Gap reduction.} Results show how visual grounding affects the multilingual gap. In audio-only models (\texttt{SSL$_A$}), performance drops when moving from the monolingual to the bilingual setting. The multilingual gap $y$ corresponds to a 31.5\% relative difference in ABX phonetic discrimination error rates (29.41\% within-speaker and 33.42\% across-speaker).
Models trained with visual grounding (\texttt{VGS+}) show a smaller degradation. The gap $w$ is 8.04\% (5.46\% within-speaker and 10.43\% across-speaker). This reduction from 31.5\% ($y$) to 8.04\% ($w$) supports our first hypothesis that visual grounding mitigates the performance disparity between monolingual and bilingual models ($y>w$ in Fig.~\ref{fig:method}).

\vspace{0.5em}

\noindent \textbf{(2) Differential benefit.} We next analyze how visual grounding improves performance in mono- and bilingual setups. The monolingual \texttt{SSL$_A$} model attains an ABX phonetic discrimination error rate of 7.11\% (6.46\% within-speaker, 7.75\% across-speaker), while the monolingual \texttt{VGS+} model reaches 6.34\% (5.86\% within-speaker, 6.81\% across-speaker), a relative improvement $x$ of 10.84\%. In the bilingual setting, the same comparison yields a relative improvement $z$ of 26.74\%. These results support our second hypothesis: visual grounding benefits bilingual models more than monolingual ones ($z>x$ in Fig.~\ref{fig:method}).

\vspace{0.5em}

Both patterns ($y>w$ and $z>x$ in Fig.~\ref{fig:method}) also hold when changing the latent layer used for extraction, when selecting the best ABX score across layers, and when running ABX on clustered units obtained with $k$-means ($k{=}50$) following \cite{nguyen2020, deseyssel2023}, rather than on the continuous representations.

\vspace{0.5em}

\noindent \textbf{Does visual grounding improve language discrimination?}
The reduction of the phonetic multilingual gap raises the question of why visual grounding helps. One possible factor is improved language discrimination, which would reduce cross-lingual interference. We therefore evaluate models on the ABX language discrimination task (Table~\ref{tab:langdisc}). In the audio-only setting, bilingual models show higher error than monolingual ones, reflecting cross-lingual interference. By contrast, visually grounded models (\texttt{VGS+}) show substantially lower error rates, both monolingual (22.98\%) and bilingual (33.69\%), indicating that vision strengthens the separation between languages. These results are consistent with the idea that visual grounding acts as an interlingual bridge: it helps bilingual models keep languages apart internally, thereby reducing cross-lingual interference and contributing to the smaller phonetic multilingual gap. However, further work is needed to establish whether this is a direct causal effect.

\begin{table}[t]
\small
\centering
\begin{tabular}{llr}
\hline
Model & Dataset & ABX LangDisc (\%) $\downarrow$ \\
\hline
\texttt{SSL}      & Monolingual 1k &  34.04 \\
\texttt{SSL}      & Bilingual 1k               & 36.66 \\
% \texttt{SSL}      & Bilingual 2k               & 31.56 \\
\hline
\texttt{SSL$_A$}  & Monolingual    & 37.16 \\
\texttt{SSL$_A$}  & Bilingual 1k               & 39.93 \\
\hline
\texttt{VGS+}     & Monolingual     & 22.98 \\
\texttt{VGS+}     & Bilingual 1k               & 33.69 \\
\hline
\end{tabular}
\caption{ABX language discrimination error rates (chance level: 50\%). Monolingual results are averaged over English and French. LL: LibriLight, AC: Audiocite, ML: ML-COCO.}
\label{tab:langdisc}
\end{table}

\vspace{0.5em}
\noindent \textbf{Robustness to domain shift.}
Complementing these findings, comparing the first stage (\texttt{SSL}) with the second stage (\texttt{SSL$_A$} and \texttt{VGS+}) reveals a consistent pattern across stages. Incorporating ML-COCO data causes \texttt{SSL$_A$} models to degrade across both monolingual and bilingual settings on both phonetic and language discrimination tasks, whereas \texttt{VGS+} models consistently improve. Given the mismatch between the read speech in stage one and the synthesized speech in ML-COCO, visual grounding appears to yield representations more robust to domain mismatch, consistent with prior findings \cite{shi2022}.

\section{Conclusion}
\label{sec:conclusion}

Our experiments show that even limited visual grounding can substantially reduce the multilingual gap in bilingual SSL models. Rather than doubling the training hours to cover both languages, bilingual models with visual grounding already approach monolingual performance while using only half the data per language. Like bilingual infants who use multimodal cues to cope with reduced exposure, visually grounded SSL models can compensate for limited per-language data in multilingual training. This highlights visual grounding as a data-efficient alternative to purely data-hungry strategies. Our results also corroborate prior work showing that visually grounded speech models yield representations more robust to domain shifts \cite{shi2022}, strengthening the case for visual grounding as a general strategy for multilingual SSL.

One limitation of our work is the use of fixed weights for the dual objectives in FaST-VGS+, motivating exploration of alternative weightings, but also different integration strategies of visual grounding. Other open questions include how little audio-visual data is sufficient, and whether improvements in discrimination directly cause reductions in the multilingual gap or are merely correlated. Finally, broader generalization should be assessed by extending beyond the bilingual setting to larger multilingual scenarios, examining language pairs with different phonetic overlap, and testing whether gains in zero-shot ABX discrimination transfer to downstream tasks.

% A limitation of our work is the use of fixed weights for the dual objectives in FaST-VGS+. Future work should test alternative weightings and strategies for incorporating visual grounding throughout training rather than only at the end. Other open questions include how little audio-visual data is sufficient, and whether improvements in discrimination directly cause reductions in the multilingual gap or are merely correlated. Finally, broader generalization needs to be assessed, both by extending beyond the bilingual setting to larger multilingual scenarios, by examining language pairs with different phonetic overlap, and by testing whether the gains observed in zero-shot ABX discrimination transfer to downstream tasks.
\newpage % ADDED due to small size of references it looks ackward if left in the same page as the conclusion (title fontsize don't match)
\small  % ADDED
\bibliographystyle{IEEEbib}
\bibliography{strings,refs}

\end{document}